%% file: templateArxiv.tex
\documentclass{article}

\usepackage{PRIMEarxiv}

\usepackage[utf8]{inputenc} % allow utf-8 input
\usepackage[T1]{fontenc}    % use 8-bit T1 fonts
\usepackage{hyperref}       % hyperlinks
\usepackage{url}            % simple URL typesetting
\usepackage{booktabs}       % professional-quality tables
\usepackage{amsfonts}       % blackboard math symbols
\usepackage{nicefrac}       % compact symbols for 1/2, etc.
\usepackage{microtype}      % microtypography
\usepackage{lipsum}
\usepackage{fancyhdr}       % header
\usepackage{graphicx}       % graphics
\graphicspath{{media/}}     % organize your images and other figures under media/ folder

\usepackage{cite}
\usepackage{amsmath,amssymb,amsfonts}
\usepackage{algorithmic}
\usepackage{graphicx}
\usepackage{textcomp}
\usepackage{makecell}
\usepackage{balance}
\usepackage{bbm}

\makeatletter
\let\NAT@parse\undefined
\makeatother
\usepackage{hyperref}

\hypersetup{
  colorlinks,
  citecolor=blue,
  linkcolor=blue,
  urlcolor=blue}

\newcommand\norm[1]{\left\lVert#1\right\rVert}

\title{RadGazeGen: Radiomics and Gaze-guided Medical Image Generation using Diffusion Models
}

\author{
  Moinak Bhattacharya \\
  Dept. of Biomedical Informatics \\
  Stony Brook University \\
  \texttt{\{moinak.bhattacharya@stonybrook.edu} \\
  \And
  Gagandeep Singh \\
  Dept. of Radiology \\
  Colombia University \\
  \texttt{gs3202@cumc.columbia.edu} \\
  \And
  Shubham Jain \\
  Dept. of Computer Science \\
  Stony Brook University \\
  \texttt{jain@cs.stonybrook.edu} \\
   \And
  Prateek Prasanna \\
  Dept. of Biomedical Informatics \\
  Stony Brook University \\
\texttt{prateek.prasanna@stonybrook.edu} \\
}

\begin{document}
\maketitle

\begin{abstract}
In this work, we present \textit{RadGazeGen}, a novel framework for integrating experts' eye gaze patterns and radiomic feature maps as controls to text-to-image diffusion models for high fidelity medical image generation. Despite the recent success of text-to-image diffusion models, text descriptions are often found to be inadequate and fail to convey detailed disease-specific information to these models to generate clinically accurate images. The anatomy, disease texture patterns, and location of the disease are extremely important to generate realistic images; moreover the fidelity of image generation can have significant implications in downstream tasks involving disease diagnosis or treatment repose assessment. Hence, there is a growing need to carefully define the controls used in diffusion models for medical image generation. Eye gaze patterns of radiologists are important visuo-cognitive information, indicative of subtle disease patterns and spatial location. Radiomic features further provide important subvisual cues regarding disease phenotype. In this work, we propose to use these gaze patterns in combination with standard radiomics descriptors, as controls, to generate anatomically correct and disease-aware medical images. \textit{RadGazeGen} is evaluated for image generation quality and diversity on the REFLACX dataset 
% (n=110)
. To demonstrate clinical applicability, we also show classification performance on the generated images from the CheXpert test set (n=500) and long-tailed learning performance on the MIMIC-CXR-LT test set (n=23550). 
% The code will be released upon acceptance.
\end{abstract}

% keywords can be removed

\section{Introduction}

Current clinical applications of AI encounter significant hurdles such as data scarcity, missing modalities, and dataset imbalances, highlighting the importance for enhancing the generation of high-quality medical images to bolster automated patient diagnosis, prognosis, and to expedite clinical research. In the last few years, diffusion models~\cite{ho2020denoising,nichol2021improved} have gained prominence in image generation. More recently, these models have made remarkable progress in conditional image generation, mainly in text-to-image (T2I) generation~\cite{rombach2022high}. 
% \mb{Other recent applications include video generation\cite{}, multi-modal \mb{image} generation\cite{}, etc.} 
These models take text descriptions or captions as input and attempt to synthesize images that have maximum resemblance with the provided description. 

Recent T2I models trained on huge amounts of data have achieved state-of-the-art performance~\cite{gao2023masked}. However, this type of textual conditioning has several limitations. In the case of complex textual structuring i.e., when the text has a description of complex objects or ill-defined contents, the T2I diffusion models show poor image generation capabilities~\cite{mehrabi2023resolving}. And vice versa, to generate a complex image, the text description design is often ill-defined. Even in the case of a well-defined text description, the generated images from T2I models occasionally have missing objects, incorrect shapes, color, position, etc.~\cite{ghosh2024geneval}. This problem stems from the fact that there is an inherent ambiguity of obviousness between text description and real-life images. A simple example is, \textit{an elephant and a bird flying} intuitively means that only the bird is flying. However, the ambiguity can be that in an unrealistic scenario both the elephant and the bird are flying. Hence, instead of generating an image where there is an elephant on the ground and a bird in the sky, the generated image shows both the elephant and bird flying\cite{mehrabi2023resolving}.  This particular problem is in fact more common in the case of medical images. For example, if the radiologist text indicates that there is a presence of a certain abnormality, \textit{enlarged cardiac mediastinum} but it is not very specific about the location and severity, the generated CXR can have enlargement is any direction and may have less severe enlargement than anticipated. Hence, location and severity are critical to clinically accurate image generation. Even after several iterations of prompt design, the exact desired image is extremely difficult to generate. Another important problem is that radiology reports contain common disease related diagnostic information and often exclude any global anatomical or patient demographic information. Moreover, description of local changes in image patterns, and their relationship to the disease, are also not present in such reports. This limitation is a real challenge to medical image generation. This motivates us to explore other clinically-relevant conditioning approaches.

\begin{figure*}[t]
\centering
\includegraphics[height=7.5cm]{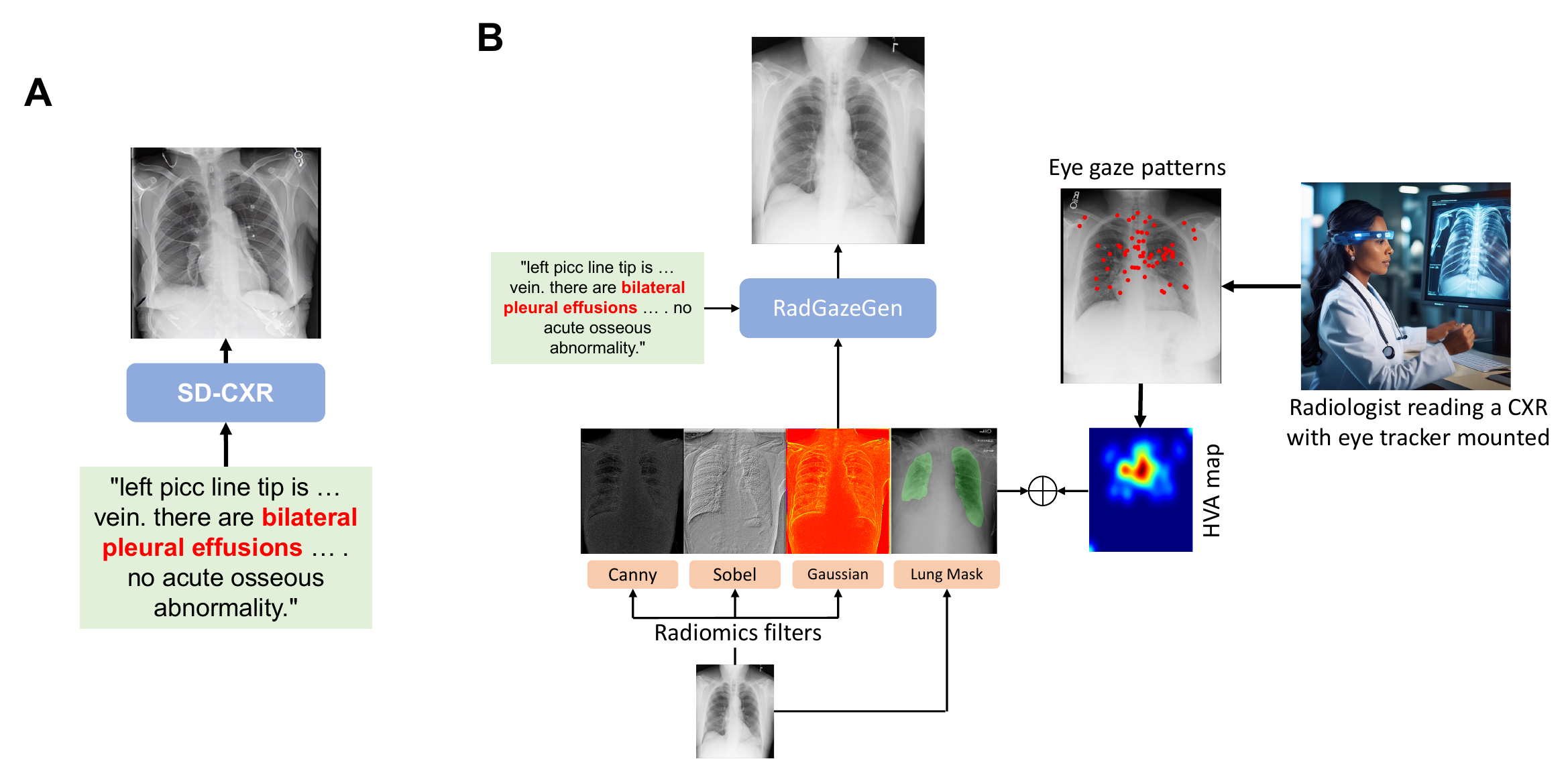}
\caption{\textbf{A.} Baseline methods add generic controls (for example text) to the diffusion models for medical image generation, \textbf{B.} Our method, \textit{RadGazeGen}, combines different radiomics filter maps and lung segmentation mask with radiologist's eye gaze patterns, and uses these controls for clinically accurate medical image generation.}
\label{fig:teaser}
\end{figure*}

% Although, T2I diffusion models generate realistic images based on a text description, synthesizing images with the desired pose, shape, depth, etc., is difficult from just text descriptions.
Generating images with the desired structure, texture, and attribute location may require several trials to design an unambiguous and clearly instructed prompt. To address this, additional controls~\cite{zhang2023adding} are often fed to diffusion models to generate the desired structure. These controls are fine-grained information that contain required structural, textural, and positional features to generate specific desired images. Images of edge maps, depth maps, pose skeletons, segmentation maps, can all be used as controls. 
% From these images, using image-to-image translation models~\cite{}, the structure, texture, or position is mapped to generate images. 
Recently, remarkable progress has been made in this domain where more fine-grained additional controls like spatial maps~\cite{couairon2023zero} have demonstrated superior image generation capabilities. However, in the case of medical image generation, these controls are not well-defined and the field is unexplored. 
% \pp{Radiomics comes out of nowhere. You need to set the stage by saying what are kinds of controls have been used for med image generation tasks - why there's a need for better controls - how radiomics capture disease properties - and hence these may be used as controls. But I think before even going into radiomics, you should mention eye gaze as controls first. Thats the main contribution.}. 

Medical image generation is a complex task because apart from generating high quality images, the quality of generated anatomical structure, and more importantly the precise representation of disease patterns are extremely critical. Current text-conditioned diffusion models fall short in effectively addressing these critical aspects of medical image generation. In medical imaging, radiomics features have long been used for various tasks involving disease diagnosis, prognosis, and treatment response assessment. Radiomics features are image transformations that capture a variety of properties of medical images pertaining to local and global changes in intensity distributions - in this process, the textural patterns of disease manifestation are quantified. In this work, \emph{we hypothesize that these different clinically-relevant radiomic feature maps can be used as specialized controls for accurate medical image generation.}

In clinical domains such as radiology and pathology, experts' visual patterns have been shown to have an impact on medical image interpretation~\cite{bhattacharya2022radiotransformer, bhattacharya2022gazeradar,bhattacharya2024gazediff}. These eye gaze patterns harbor contextually rich visuo-cognitive data related to the occurrence and patterns of diseases~\cite{just1976eye}.
% An expert while scanning a medical image searches for disease features in specific regions based on their years of medical training and experience. Hence, over the years, the patterns of scanning evolve and these become an important source of information in understanding the expert's diagnostic behaviour. In the last few years, medical image diagnosis using deep learning models has gained prominence. Several models have been proposed that have achieved state-of-the-art performance for different medical tasks. The main drawback of a deep learning model is that it tends to look at non-disease relevant regions (often termed as 'shortcuts') to make a diagnosis. This raises a serious question on clinical relevance of these deep learning models. 
Recently, several methods have shown that integrating gaze patterns into deep learning frameworks helps improve the model performance for several diagnostic tasks~\cite{bhattacharya2022gazeradar,bhattacharya2022radiotransformer}. 
% By integrating these eye gaze patterns into the deep learning models, the models tend to look at regions where a radiologist might look during a diagnosis. 
\emph{Our next hypothesis is that the human visual attention (HVA) maps computed from these eye gaze patterns can be used as relevant controls to translate the disease occurrence and visual patterns into a diffusion model.}
In this work, we propose to integrate these eye gaze patterns into diffusion models for medical image generation. Previous works have explored medical image generation from text conditions where disease patterns generated can be unrealistic and not located in the desired regions. By integrating these gaze patterns, we aim to generate medical images with not just realistic disease patterns but ones where such patterns are in the regions where an expert is focusing during diagnostic interpretation. 

\textbf{Summary of Motivation.} 
% The motivation of the paper originates from generating gaze-guided CXR images. 
Owing to the ambiguity between the radiology text description and images, diffusion models face difficulty in generating realistic images with accurate disease patterns. In recent works in the non-medical domain, the addition of spatial features as controls improves T2I generation capabilities. In medical image generation, by adding such controls, anatomical structure of the CXRs can be accurately generated; however, translating the disease patterns 
% from  
still remains a challenging problem\cite{pinaya2023generative}. Our work is based on the premise that, to translate the disease patterns and locations, the radiologists' eye gaze fixations and radiomic feature maps can be used as clinically-relevant controls. 

In this work, the radiomics filter maps are combined with HVA maps and provided as controls to diffusion models. However, composing these filters as a single control is difficult; hence,  we propose separate frameworks for radiomics filters maps and HVA maps. For training diffusion models with the radiomics filters maps, we design Rad-CN and for training diffusion models with the HVA maps, we design HVA-CN. The combined Rad-CN and HVA-CN make our \textit{RadGazeGen} architecture. We demonstrate the efficacy of our approach for several tasks. 
% We compare the quality of the generated images with several diffusion models and 
\textit{RadGazeGen} outperforms other diffusion baselines, thereby highlighting the importance of radiomics filters and HVA maps as controls to improve clinical diagnostic quality of the generated images. Furthermore, we use a pretrained disease classification model for inference on the generated images. We show that our generated images contains the disease patterns as in the training dataset distribution and achieve comparable AUCs on different disease pathologies; this is further qualitatively corroborated by our collaborating radiologists. Finally we also show results on a Long-Tailed disease classification task where we generate additional CXRs of the medium and tail classes and train the classifier on the re-balanced data distribution. We observe that injection of our generated images improve the balanced accuracy and other metrics when compared to the injection of baseline generated images to the data distribution.

\begin{figure*}[t]
\centering
\includegraphics[height=8cm]{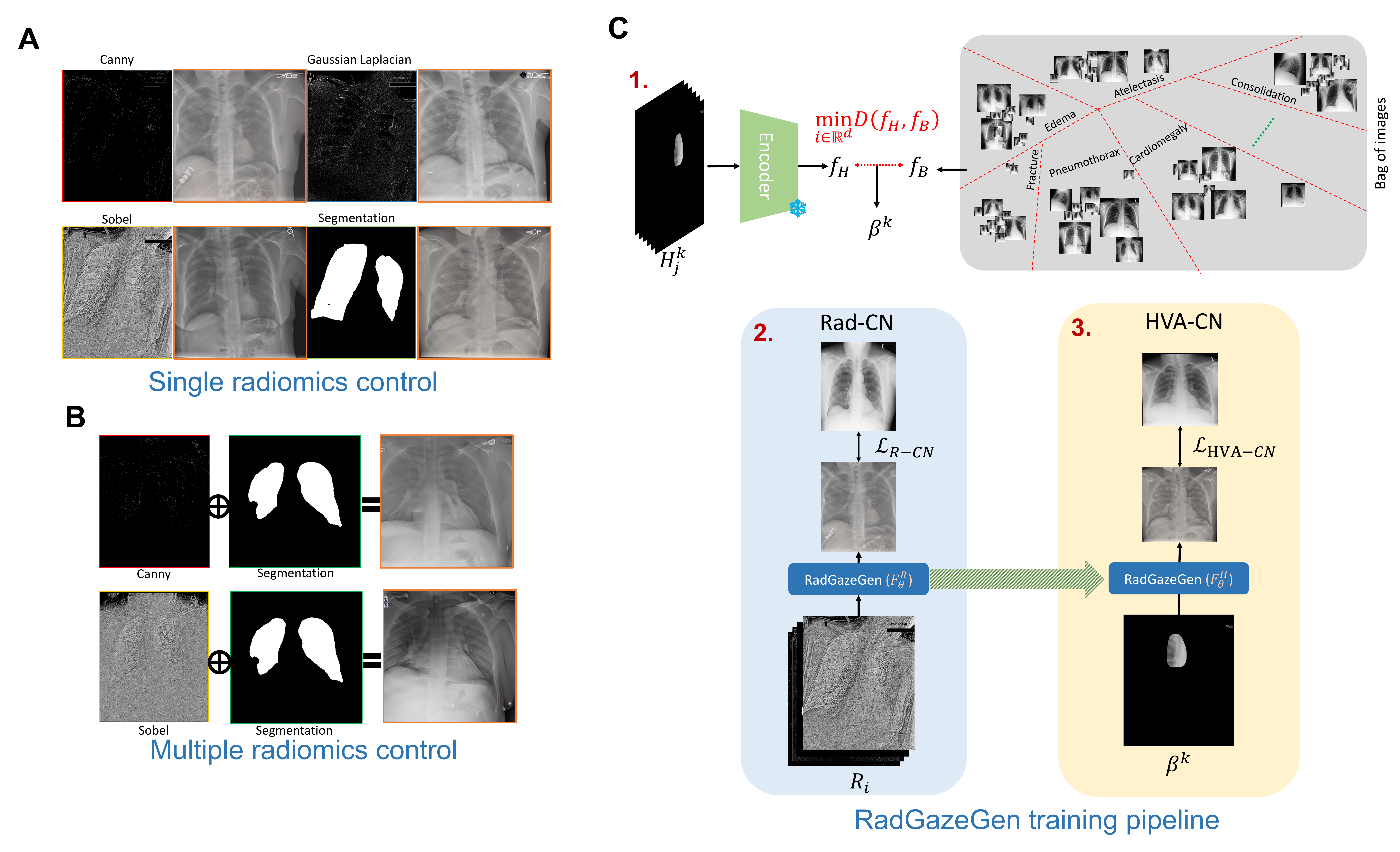}
\caption{\textbf{Overview of \textit{RadGazeGen} pipeline.} \textbf{A.} Generated CXRs for different radiomics filter maps and lung segmentation mask as controls, \textbf{B.} Generated CXRs when more than one controls are fused, and \textbf{C.} Different components of Rad-CN and HVA-CN.}
\label{fig:main}
\end{figure*}

\textbf{Contributions} The primary contributions of this work are as follows:
\begin{itemize}
    \item We propose a gaze-guided diffusion model \textit{RadGazeGen} to generate disease pattern-preserving CXR images. This model comprises of two modules, \textit{Rad-CN} takes radiomic 
    % filters 
    feature maps as inputs to generate anatomically and texturally-aware CXR images and \textit{HVA-CN} which takes radiologists' eye gaze patterns and feeds the same to the \textit{Rad-CN} to generate disease pattern and location-aware CXR images.
    \item We demonstrate the performance of \textit{RadGazeGen} for medical image classification and long-tailed classification
    % learning 
    tasks. 
    % We show that \mb{}
    % \item We also propose a dataset with samples for the tail classes for NIH-CXR-LT and MIMIC-CXR-LT datasets.
\end{itemize}
In Figure \ref{fig:teaser}, we show our \textit{RadGazeGen} pipeline and compare it with existing methods. \textit{RadGazeGen} takes both radiomics filter maps and Human Visual Attention (HVA) maps, that are computed from eye gaze patterns of radiologists, as control inputs and generates a CXR image.

\section{Related Works}
% \subsection{Eye-gaze in Medical Image Analysis}

\begin{figure*}[t]
\centering
\includegraphics[height=7.5cm]{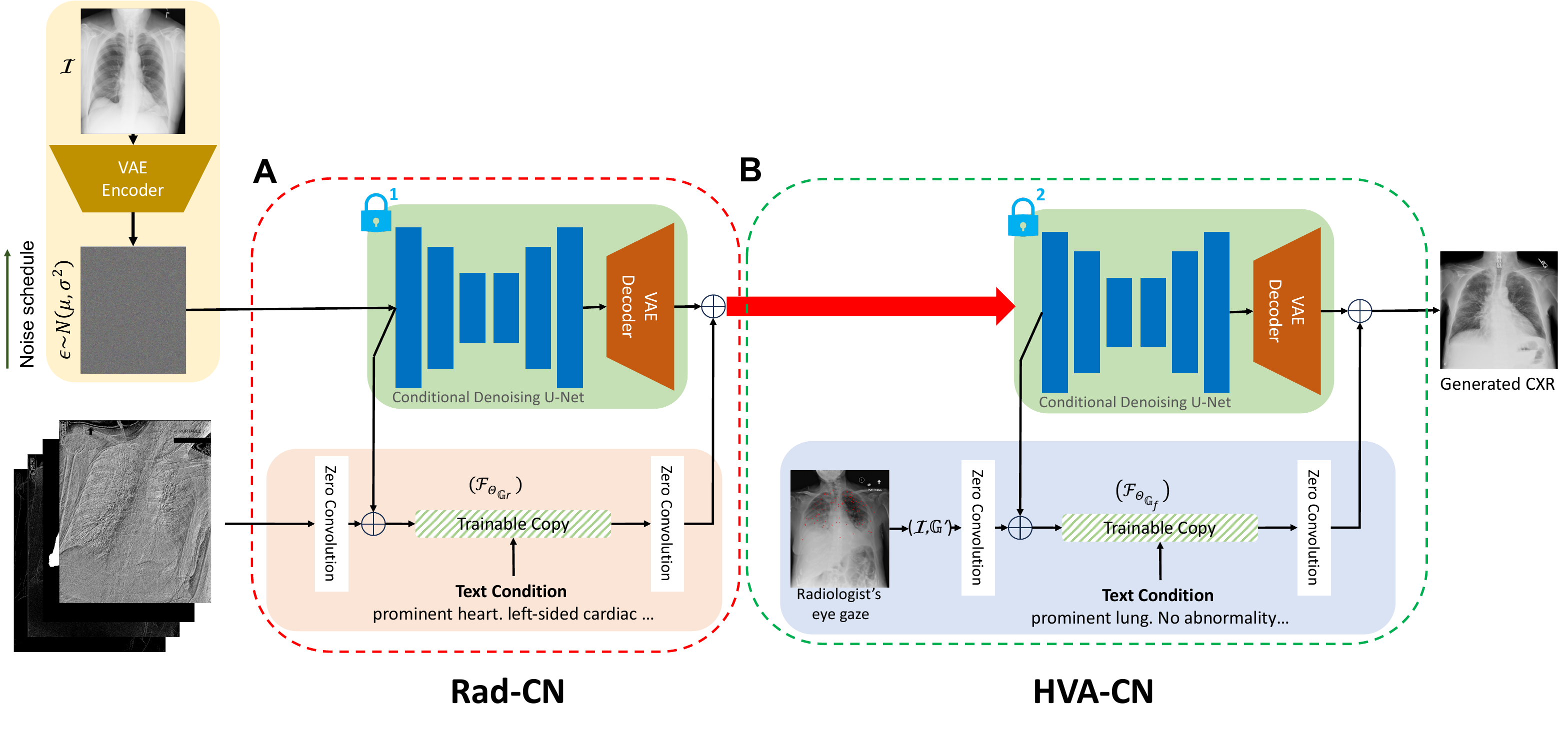}
\caption{\textbf{Overview of the Rad-CN and HVA-CN architecture.} The pre-trained SD model is locked and the Rad-CN (shown in \textbf{A}) and HVA-CN (shown in \textbf{B}) are finetuned by making a trainable copy of the SD model.}
\label{fig:architecture}
\end{figure*}

\textbf{Eye-gaze in Medical Image Analysis.}
The eye gaze of experts has demonstrated immense potential in several medical applications like disease classification\cite{bhattacharya2022gazeradar,bhattacharya2022radiotransformer} and segmentation\cite{wang2023gazesam}, for various modalities, including, radiology\cite{stember2020integrating}, pathology\cite{sudin2022digital}, retinopathy\cite{clark2019potential}, and ECG\cite{sqalli2022understanding,tahri2021interpretation}
% Eye tracking is used as a diagnostic tool in neurological diseases like Parkinson’s disease\cite{bek2020measuring}, Alzheimer’s disease\cite{zammarchi2021application}, Attention-Deficit/Hyperactivity Disorder\cite{lev2022eye}, etc. 
After successfully leveraging eye gaze patterns in general computer vision tasks like object detection\cite{smith2013gaze,cho2021human}, image segmentation\cite{amrouche2018activity,james2007eye,shi2017gaze}, action localization\cite{shapovalova2013action,mathe2014actions}, and activity recognition \cite{courtemanche2011activity,min2021integrating}, recent applications have ventured into utilizing them in the clinical domain. Understanding gaze patterns of medical professionals, such as radiologists/pathologists, and providing this information as an auxiliary signal to ML models can potentially reduce misdiagnosis\cite{roshan2023eye}. This claim was validated by recent works\cite{bhattacharya2022gazeradar,bhattacharya2022radiotransformer} which employed teacher-student knowledge distillation models to improve disease diagnosis.
% Integrating eye gaze in clinical use is extremely important for understanding and improving diagnostic patterns\cite{}. Radiologists often suffer from strain and fatigue and that leads to clinical errors\cite{} and studies show that by using eye-gaze patterns changes of misdiagnosis can be reduced\cite{}.
% Another potentially important application of gaze patterns is in training of residents and medical students\cite{}. 
In this work, we propose a new framework of diffusion models that generate gaze-guided clinically relevant medical images.

% .... improving diagnostic performance....\\
% Over the last few years, eye gaze patterns have been successfully used with deep learning frameworks like convolutional neural networks\cite{}, graph neural networks\cite{wang2024gazegnn}, vision transformers\cite{ma2023eye}, etc.  However, radiologists' eye gaze in medical image generation has immense clinical usage which remains unexplored. In this work, we propose a new framework of diffusion models that generates gaze-guided clinically relevant medical images.\\
% \subsection{Text-to-Image Diffusion}
 \textbf{Text-to-Image Diffusion.}
Since their introduction\cite{sohl2015deep}, Diffusion models have been applied extensively to image generation tasks\cite{ho2020denoising,nichol2021improved}. Latent diffusion models (LDMs) deal with latent spaces of the images to perform the diffusion steps\cite{rombach2022high}. Text-to-image diffusion models integrate text embeddings with image latent space embeddings using CLIP\cite{radford2021learning}, achieving state-of-the-art performance in image generation tasks. 
% Stable Diffusion\cite{nichol2021improved,ho2020denoising} is an ... advancement of LDMs. 
Recent commercial applications of text-to-image generations are DALL-E2\cite{dalle2,ramesh2022hierarchical} and Midjourney\cite{midjourney}.
Recently, LDMs have been used for different medical imaging modalities for a wide range of applications  such as image-to-image translation\cite{ozbey2023unsupervised,durrer2023diffusion}, reconstruction\cite{peng2023one,chung2022improving,peng2022towards,xie2022measurement,chung2022come,song2021solving,chung2022score,jalal2021robust}, denoising\cite{xiang2023ddm,gong2023pet}, registration\cite{kim2021diffusemorph,qin2023fsdiffreg}, segmentation\cite{wang2024towards,laousy2023certification,carrion2023fedd,rahman2023ambiguous,bieder2023diffusion,zbinden2303stochastic}, classification\cite{yang2023diffmic,ijishakin2023interpretable,hardy2023improving}, etc. 
Despite their overall success, the ambiguity between text descriptions and images in text-to-image diffusion models can severely impact the quality of generated images and often lead to suboptimal generations\cite{mehrabi2023resolving}.
% Although diffusion models can generate good-quality medical images, in text-to-image diffusion models, the ambiguity between text descriptions and images often leads to suboptimal generations\cite{}. 
% This ambiguity leads the models to generate clinically irrelevant images. 
% To address this, recent methods propose prompt engineering to improve the quality of the generated images\cite{}. 
% In this work, we use a SD model as base and train it with MIMIC-CXR dataset for downstream medical image generation tasks.

% \subsection{Controllable Diffusion Models}
\textbf{Controllable Diffusion Models}
Recently, diffusion models in which one or more controls are used are gaining traction. These controls help in generating custom and task-specific images. The controls like color variation\cite{meng2021sdedit}, and inpainting\cite{avrahami2022blended,ramesh2022hierarchical} can be directly provided by the image diffusion process. Controllable diffusion models can be broadly classified into two categories: models trained from scratch\cite{huang2023composer} and finetuned 
lightweight adapters that are finetuned by freezing the pretrained T2I diffusion models\cite{zhang2023adding,mou2023t2i}. Composer\cite{huang2023composer} uses single or multiple conditions to generate images by training large diffusion models. T2I-Adapter\cite{mou2023t2i}, GLIGEN\cite{li2023gligen} and ControlNet\cite{zhang2023adding}. finetune an adapter from frozen large pretrained diffusion models. 
% This methods helps in reducing training cost and facilitates ease of research. 
More recently, several methods like UniControl\cite{qin2023unicontrol} and UniControlNet\cite{zhao2024uni}, have been proposed where multiple controls are unified to generate more specific images with different styles and contents. CoAdapter\cite{mou2023t2i} and Multi-ControlNet\cite{zhang2023adding} are variations of T2I-Adapter and ControlNet, respectively, where models trained with different conditions are fused to generate images from composite conditions. However, for medical image generation, the use of additional controls is still in its infancy.
% defined. 
% \pp{Why has this not been done - any challenges?} 
Given the diversity of modalities and downstream tasks in medical images, the design of controls in this domain depends largely on the specific modality and task.
% \mb{There are numerous modalities in medical images and also there are a variety of downstream tasks. Hence, the design of controls depends largely on the kind of modality and downstream task.} 
Medical image translation\cite{pinaya2023generative} is an important field of research that has immense clinical importance with applications ranging from generating multi-modality images to data augmentations. However, this domain of image translation using diffusion models with additional controls is relatively unexplored.

\textbf{Radiomics.} Radiomics encompasses the computerized extraction of quantifiable data from radiological images, typically subtle radiographical features. These features, which capture morphological and textural attributes, have been shown to be effective in several clinical applications involving radiographs~\cite{han2021pneumonia, tamal2021integrated}. The role of such clinically relevant descriptors in generative models is unexplored.

In summary, there is no existing work that studies clinically relevant controls for radiology image generation. To address this limitation, we propose \textit{RadGazeGen}, which uses radiomics filter maps and radiologists' eye gaze patterns as controls to generate clinically accurate medical images. 

\section{Method}
An overview of the proposed method is shown in Figure~\ref{fig:main}. In this work, we first train individual ControlNet (CN) models with different radiomics maps and segmentation masks as a control (shown in A). Then, we perform zero-shot inference by combining two or more of these controls (shown in B). The proposed method \textit{RadGazeGen} comprises two main modules, namely, \textit{Radiomics ControlNet (Rad-CN)} and \textit{Human Visual Attention ControlNet (or HVA-CN)}. For \textit{Rad-CN} training, the radiomics maps and the lung segmentation mask are combined to form unified controls. HVA maps are computed from the eye gaze patterns of radiologists. For \textit{HVA-CN} training, these HVA maps in a specified manner are fed as controls to the trained \textit{Rad-CN}.

In this section, we start by introducing the preliminary concepts that are used in developing the methodology of the paper (\ref{preliminary}). In \ref{gazecn}, we discuss the different components of the proposed methodology \textit{RadGazeGen}. The primary components of \textit{RadGazeGen}, \textit{Rad-CN} is discussed in  \ref{radcn} and \textit{HVA-CN} is discussed in  \ref{hvacn}. Finally, we discuss the inference of \textit{RadGazeGen} in \ref{inference}.
% Finally, we discuss the applications of \textit{GazeControlNet} in medical image analysis mainly for thoracic diseases classification and long-tailed learning(Subsection \ref{longtailed}).

\subsection{Preliminary}
\label{preliminary}
% In this work, we discuss several advanced concepts, and the basics for these concepts are discussed in this subsection. 
% \mb{Here, we discuss the basics for the concepts discussed in this section}. 
Diffusion probabilistic models \cite{ho2020denoising}, often referred to as diffusion models, are generative models that are parameterized forms of Markov chains trained using variational inference. Let us consider an image input $x$, the forward process is defined as adding Gaussian noise $\epsilon\sim\mathcal{N}(\mu, \sigma^2)$ to $x$ according to a variance schedule $\beta_t\in\{\beta_0, \beta_1, ..., \beta_T\}$, where $T$ is the number of total time steps.  
% ...
\begin{equation}
    % \begin{aligned}
        q(x_{1:T}|x_0) := \prod_{t=1}^T q(x_t|x_{t-1}), q(x_t|x_{t-1}) \\ := \mathcal{N}(x_t; \sqrt{1-\beta_t}x_{t-1}, \beta I)
    % \end{aligned}
\end{equation}
and the reverse process is defined as,
\begin{equation}
    p_\theta(x_{0:T}):=p(x_T)\prod_{t=1}^T(x_{t-1}|x_t).
\end{equation}
The training is done using variational lower bound (ELBO), shown as,
\begin{equation}
    \mathbb{E}[-\log p_\theta(x_0)]\leq\mathbb{E}_q\left[-\frac{p_\theta(x_{0:T})}{q(x_{1:T}|x_0)}\right]
\end{equation}
In our case, we train the Stable Diffusion (SD) model with chest X-rays (CXRs) as image $x$, and the radiologist's reports as text conditioning, $c$, in a T2I diffusion manner. Radiologists eye gaze patterns, represented as $\mathbb{G}$, are collected when they scan a CXR during diagnosis. The radiologists also provide text transcripts after diagnosis. We compute HVA maps from $\mathbb{G}$, similar to the method described in \cite{bhattacharya2022radiotransformer} (discussed in detail in Supplementary). 
In short, the HVA maps are heatmaps computed by adding Gaussian blur to $\mathbb{G}$.

\subsection{RadGazeGen} 
\label{gazecn}
An overall pipeline of \textit{RadGazeGen} is shown in Figure \ref{fig:main}. We first train a SD model with CXRs and radiology reports as text conditions, $c_{sd}$. By doing this, the SD model is generalized for CXR images (represented as SD-CXR) and can be used for further finetuning to serve different medical image generation tasks. 
% The experiment design and dataset details are provided in Section \ref{experiments_and_results}.
Details of Rad-CN and HVA-CN architecture are shown in Figure \ref{fig:architecture}.
Next, we discuss the two major components of \textit{RadGazeGen} as follows.
\subsubsection{Radiomics ControlNet.}
\label{radcn}
The \textit{Rad-CN} module is used to finetune the SD-CXR with $n$ different radiomics feature maps, $\mathcal{R}_i$, to generate radiomics-guided CXR images
% generation 
with a text condition $c_{\mathcal{R}}$. Here, $\mathcal{R}_i\in\{\mathcal{R}_1, ..., \mathcal{R}_n\}$  are computed by applying different radiomic filters to the CXR image $\mathcal{I}^\mathcal{R}$. The objective of \textit{Rad-CN} is to translate from $\bigcup_i\mathcal{R}_i$ to real images $x$ based on $c_{\mathcal{R}}$. In Figure \ref{fig:main}.C.2, we show the different $\mathcal{R}_i$ is fed to $\mathcal{F}^{\mathcal{R}}_\theta$ which generates $\hat{\mathcal{I}}^\mathcal{R}$. The training loss is defined as
\begin{equation}
    \mathcal{L}_{R-CN}(\theta_\mathcal{R}) = \mathbb{E}_{z^\mathcal{R}, \epsilon(x), t, c_\mathcal{R}, \mathcal{R}_i}\left[\norm{\epsilon-\epsilon_{\theta_\mathcal{R}}(z^\mathcal{R}_t, t,c_\mathcal{R}, \mathcal{R}_i)}^2_2\right]
\end{equation} 
where $t$ is time step, $z^\mathcal{R}_t$ is the noise at each time step $t$, $\theta_\mathcal{R}$ is the trainable parameter of Rad-CN.
\subsubsection{Human Visual Attention ControlNet.}
\begin{figure}[t]
\centering
\includegraphics[height=5.5cm]{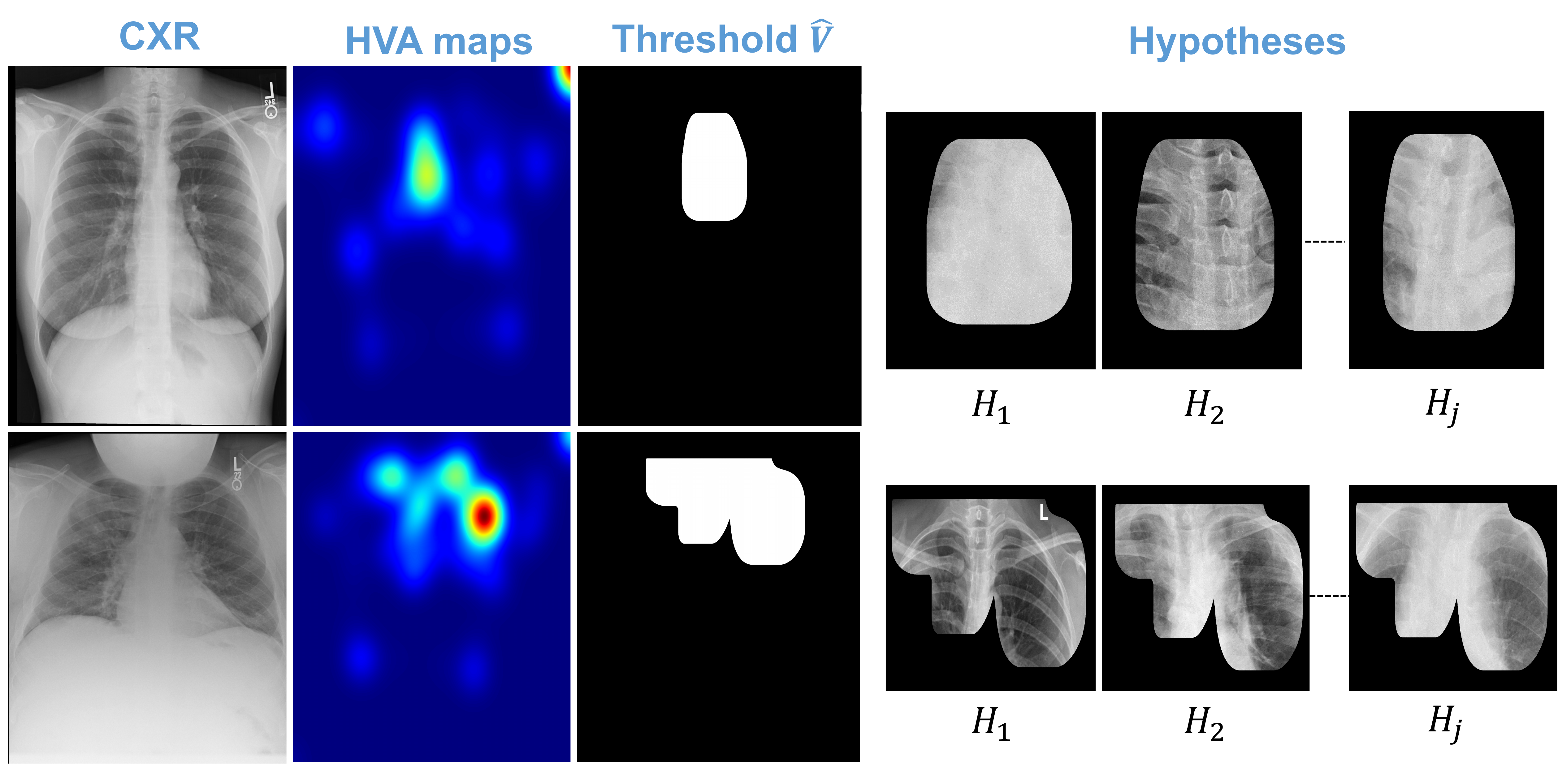}
\caption{\textbf{Hypotheses generation.} HVA maps are thresholded to create masks, which are then mapped to various disease pathologies to generate hypotheses.}
\label{fig:hypotheses}
\end{figure}
\label{hvacn}Here, we discuss how eye gaze patterns are used as a control for CXR generation. \textit{HVA-CN} (shown in Figure \ref{fig:main}.C.3) has two major components, gaze-guided hypotheses generation and training based on the best hypotheses generated. First, we compute HVA maps, $\mathcal{V}^m$ from $\mathbb{G}$, shown in Figure \ref{fig:hypotheses}. Here, $m$ are the different disease types. These HVA maps are then used to generate multiple  CXR hypotheses images. We then select the best hypothesis and use that as the gaze-guided control for \textit{HVA-CN} training. A detailed discussion of these components is provided next.

\textbf{Gaze-guided hypothesis generation.}
% Here, we discuss hypothesis generation based on $\mathbb{G}$. 
These hypotheses are multiple generated CXRs based on a bag of images $\mathbb{B}^K$. This bag contains CXRs of $K$ different thoracic diseases.
To generate these hypotheses, first we apply a threshold to $\mathcal{V}^m$, such that $\hat{\mathcal{V}^m}(\textbf{x}, \textbf{y})=\left\{\def\arraystretch{1.2}\begin{tabular}{@{}l@{\quad}l@{}}
  $255.$ & if $\mathcal{V}(\textbf{x}, \textbf{y})\geq\lambda$ \\
  $0.$ & otherwise.
\end{tabular}\right.$, where $\textbf{x}, \textbf{y}$ are the x- and y-coordinates of the image. Here $\hat{\mathcal{V}}$ is the image obtained after the threshold, as shown in Figure \ref{fig:hypotheses}. Then, we apply the bitwise AND operation to a set of images $\mathbb{B}_j^k$ and $\hat{\mathcal{V}}^m$ as a binary mask (Figure \ref{fig:hypotheses}), such that $k=m$. 
% ... 
Finally, we obtain the gaze-guided hypotheses, represented as $\mathcal{H}_j^k$.
From these multiple hypotheses, the best hypothesis is selected by comparing it with $\mathbb{B}^k$. The hypothesis with the maximum similarity score is finally selected, shown as $\beta^k$. This score is calculated by computing the distance between the features ofsla
$\mathbb{B}^k$, $f_\mathbb{B}$ and $\mathcal{H}^k_k$, $f_\mathcal{H}$, shown as $\mathcal{D}(f_\mathbb{B}, f_\mathcal{H})$, shown in Figure \ref{fig:main}.C.1.

\textbf{Training.}
The training of \textit{HVA-CN}, represented as $\mathcal{F}_\theta^\mathcal{V}$, as shown in Figure \ref{fig:main}.C.3
\begin{equation}
    \mathcal{L}_{HVA-CN}(\theta_\mathcal{V}) = \mathbb{E}_{z^\mathcal{V}, \epsilon(x), t, c_\mathcal{V}, \mathcal{V}}\left[\norm{\epsilon-\epsilon_{\theta_\mathcal{V}}(z^\mathcal{V}_t, t,c_\mathcal{V}, \mathcal{V})}^2_2\right]
\end{equation}
where $t$ is time step, $z^\mathcal{V}_t$ is the noise at each time step $t$, $\theta_\mathcal{V}$ is the trainable parameter of HVA-CN.
% \subsection{Downstream tasks}
% \subsubsection{Classification}
% \label{classification}
% \subsubsection{Long-tailed classification}
% \label{longtailed}
% \subsection{Training}
% \subsection{Inference}
\subsubsection{Inference.}
\label{inference}
During inference, the input set $\mathcal{I}_{(\mathcal{R}_i,\mathcal{V})}:=(\textbf{x}_{\mathcal{R}_i}, \textbf{x}_\mathcal{V})$ is fed to fused $\mathcal{F}^\mathcal{R}_\theta$ and $\mathcal{F}^\mathcal{V}_\theta$, represented as $(\mathcal{F}^\mathcal{R}_\theta \oplus \mathcal{F}^\mathcal{V}_\theta)$. This 
% representation 
is our proposed \textit{RadGazeGen} architecture. The \textit{RadGazeGen} module, $\hat{\mathcal{F}}(.)$ is expressed as
\begin{equation}
\label{inference_1}
    \hat{\mathcal{F}}(\mathcal{I}_{(\mathcal{R}_i,\mathcal{V})}):=\prod_{t^*} \mathcal{F}^\mathcal{R}_{\hat{\theta}} \oplus \mathcal{F}^\mathcal{V}_{\hat{\theta}} (\mathcal{I}_{(\mathcal{R}_i,\mathcal{V})})
\end{equation}here, $\hat{\theta}$ are the frozen parameters of Rad-CN and HVA-CN and $t^*$ are the inference steps of the model. Now, we show that our model also facilitates inference using a subset of controls from the set 
$(\mathcal{I}_{(\mathcal{R}_i,\mathcal{V})})$ by using the indicator function $\mathbbm{1}(.)$, and Equation \ref{inference_1} can be rewritten as 
$\prod_{t^*} \sum_i^{\mathcal{R}_n} \mathbbm{1}(i==\mathcal{R}_i) \mathcal{F}^\mathcal{R}_{\hat{\theta}} \oplus \mathcal{F}^\mathcal{V}_{\hat{\theta}} \big(\mathcal{I}_{(\mathcal{R}_i,\mathcal{V})}\big)$.

\section{Experiments and Results}
\label{experiments_and_results}
\textbf{Datasets.} We use CXR images and free-text radiology reports
% radiologists' 
 from the MIMIC-CXR dataset for SD training. We use the eye-gaze data from publicly available
Eye Gaze Data for Chest X-rays\cite{karargyris2021creation} (n=1083) and REFLACX dataset\cite{bigolin2022reflacx} (n=2507). For evaluation, we use n=110 CXRs from the REFLACX dataset. For classification experiments, we use the CheXpert test dataset (n=500) and for long-tailed classification experiments, we use the MIMIC-CXR-LT dataset\cite{holste2022long}. This dataset is divided into the training (n=68058), validation (n=300), balanced test (n=600), and test (n=23550) set.
% More details on specific implementation is discussed in the following subsection.
\subsection{Implementation}
\label{implementation}
\textbf{SD training.} We use CXR images of size $515\times512$ from the MIMIC-CXR dataset to finetune a SD v1.5 model. We train for 50,000 steps with a learning rate of $1e-05$ and batch size of $4$. We use gradient checkpointing and mixed precision of `fp16'. The gradient accumulation steps are set to 4 and max\_grad\_norm is set to 1. Training is done on a single Quadro RTX 8000 (48 GB) GPU.

\textbf{RadGazeGen training.} To train the different components of \textit{RadGazeGen}, the two eye gaze datasets are used. 
% We split the dataset into training and test. 
For training we use n=3580 CXRs by combining the entire first dataset with n=1083 CXRs and the samples for which only single radiologist eye gaze are collected from REFLACX (n=2507). We remove the overlapping samples from the first dataset. For testing, we use the remaining n=110 from REFLACX dataset. First, we extract different radiomics filter maps using OpenCV and Sklearn libraries and lung segmentation from \cite{lung_segmentation}. We then train Rad-CN with these radiomics filters and lung masks. The HVA maps are computed as discussed in \ref{hvacn}. We train the HVA-CN with these HVA maps. For training, we use $512\times512$ image size as input and train for $10$ epochs with a batch size set to $4$ and a learning rate of $1e-5$. We use mixed precision of ‘fp16’ and the gradient accumulation steps are set to 4.
% \textbf{Classification with generated images as training set.} We generate captions of the CXRs from the NIH training set using RGRG\cite{tanida2023interactive}. Then, we extract the different radiomics filters for the CXR images and the HVA maps are extracted from RadioTransformer\cite{bhattacharya2022radiotransformer}. We then use the generated captions, radiomics filters and HVA maps to generate images from \textit{GazeControlNet}. Then, we train a DenseNet-121\cite{huang2017densely} architecture with the generated images as training and validation set. We report our inference results on the real text set of the NIH dataset.\\
% \textit{GazeControlNet} \\
% \textbf{Classification with generated images as test set.} Here, we use the pretrained DenseNet-121 from TorchXRayVision\cite{cohen2022torchxrayvision}. 
% This DenseNet-121 is trained on multiple datasets like NIH, MIMIC, PadChest, RSNA with image resolution $B\times3\times224\times224$. 

\textbf{Classification with generated images as test set.} We generate the captions for the test set of CheXpert using RGRG\cite{tanida2023interactive}. Radiomics maps are computed as described above, followed by the computation of HVA maps. The captions, radiomics filter maps, and HVA maps computed from the test set of CheXpert are used to generate CXRs using \textit{RadGazeGen}. The pretrained DenseNet-121 model from TorchXRayVision\cite{cohen2022torchxrayvision} is used to predict the classes for the generated images.

\textbf{Long-tailed (LT) classification.} For long-tailed classification, we use the MIMIC-CXR-LT dataset. We identify the medium and tail classes and randomly select 
% radiologists 
text reports for these medium and tail classes from the MIMIC-CXR dataset. Then, we randomly select sets of radiomic filter maps and HVA maps and combine them with the 
% radiologists 
texts. 
% Similar to the previous methods the HVA maps are computed from RadioTransformer\cite{bhattacharya2022radiotransformer}.
For each disease pathology in the medium and tail classes, we generate CXR images using this set as controls such that for each pathology the CXR count is n$_i$=1100, where $i$ is the different medium and tail classes. In this way, all the classes in the MIMIC-CXR-LT dataset have $\geq1000$ CXR images. A ResNet-50\cite{he2016deep} model is then trained for LT classification.
% with cross-entropy loss, a batch size of $256$ and a learning rate of $1e-4$. 
% We train for $60$ epochs with patient of $15$.
\\
\input{tables/table_1}
\input{tables/table_2}

\subsection{Quantitative Results} 
% In this subsection, we present the quantitative results for the experiments discussed above. 
In \ref{quality}, we evaluate the quality of generated images compared to original images. In \ref{classification}, we discuss the classification results on the CheXpert dataset. In \ref{long_tailed}, we discuss the long-tailed classification results on the MIMIC-CXR-LT dataset. Finally, in \ref{ablation}, we discuss the ablation analysis results. 
\subsubsection{Quality of image generation.} 
\label{quality}
First, we discuss the quality and diversity of the generated images. We compare our proposed method with different SD and ControlNet baselines. In Table \ref{tab:quantitative1}, we report the Fréchet Inception Distance (FID) scores for the generated images and compare with T2I-Adapter, ControlNet and Multi-ControlNet for different radiomic maps and HVA controls. \textit{RadGazeGen} outperforms the baselines on 4 out of 5 controls. Though T2I-Adapter achieves the best performance for canny edge filter maps as control, \textit{RadGazeGen} outperforms ControlNet and MultiControlNet on all the controls. In Table \ref{tab:quantitative2}, we use different quantitative metrics to assess the controllability. Structural Similarity Index Measure (SSIM) is used for Canny, Sobel, and GL filter maps, Mean Intersection over Union (mIoU) for segmentation maps overlap, and CLIP score is used to assess the generated content. \textit{RadGazeGen} outperforms SD and ControlNet baselines on 3 out of 5 controls. From Table \ref{tab:quantitative1} and \ref{tab:quantitative2}, we observe that ControlNet performs better than MultiControlNet on image quality and controllability. Whereas \textit{RadGazeGen} performs better than MultiControlNet when both the models are using multiple controls. These results suggest the strength of the composability of Rad-CN and HVA-CN modules for better-quality image generation.
% \textit{GazeControlNet} reports comparable performance for the Sobel filter map control with an SSIM score of $7.6e-3$ where SD,  RoentGen, ControlNet and MultiControlNet report marginally better performance of $8.0e-3$, $7.9e-3$, $7.8e-3$, and $7.7e-3$, respectively. For segmentation mask overlap, RoentGen and ControlNet outperform \textit{GazeControlNet}.

\begin{figure*}[t]
\centering
\includegraphics[height=7.5cm]{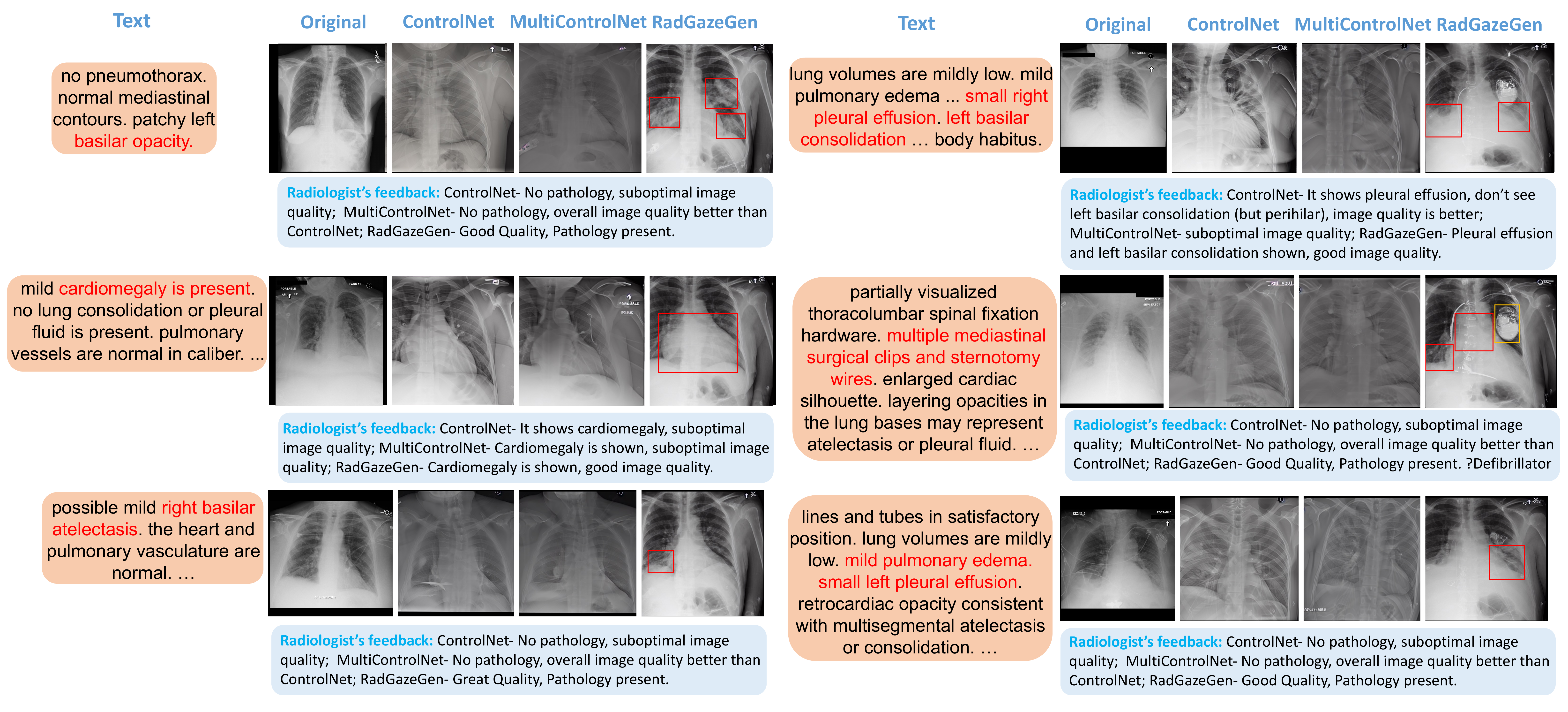}
\caption{\textbf{Qualitative comparison.} The generated CXRs for different baselines are shown along with the corresponding radiologist's text and the original image. We also show the radiologist's annotations of the disease occurrence (\textcolor{red}{red bounding box}).}
\label{fig:qualitative1}
\end{figure*}

\subsubsection{Classification Performance.}
\label{classification} The classification results on the generated images using text and other controls from the CheXpert test set are shown in Table \ref{tab:chexpert}. 
% We report the AUC for the original images in the first row. 
Disease-wise AUC for the \textit{RadGazeGen}-generated images are compared with the text-to-image diffusion model-generated images.
For context, we also provide the AUCs of the model evaluated on the real images (first row) from the CheXpert test dataset.
% those from different baselines. 
\textit{RadGazeGen} outperforms the 
% ControlNet 
text-to-image baseline on 7 out of 12 baselines, further validating the quality of the generated CXRs using the radiomics feature and HVA maps as controls. 
% SD and RoentGen models report higher AUCs when compared to the methods with radiomics and HVA maps as controls as they generate images based on just the text condition. This allows the former models to focus more on the content-based generation, whereas the latter models suffer from the issue of exact composability of different controls.
\subsubsection{Long-tailed classification performance.}
\label{long_tailed}
LT classification results on the MIMIC-CXR-LT dataset are shown in Table \ref{tab:long_tailed}. 
% We report accuracy scores for the head, medium, and tail classes and average accuracy for the entire balanced set. For the test set, we report F1 and balanced accuracy score. 
Our model is compared with ControlNet and MultiControlnet baselines.
% several baselines. Softmax is the cross-entropy loss function and Focal, LDAM, RW LDAM-DRW, and Decoupling-cRT are long-tailed losses. These baselines are trained with no additional data added to the medium and tail classes. For ControlNet, MultiControlNet, and \textit{RadGazeGen}, we add generated images to the medium and tail classes. 
% We observe that \textit{RadGazeGen} outperforms ControlNet and MultiControlNet on both the balanced test and imbalanced test sets. 
% \textit{RadGazeGen} also ourperforms Softmax on head and tail classes of the balanced test set and balanced accuracy of the test set. 
\textit{RadGazeGen} outperforms both ControlNet and MultiControlNet on head and medium and reports comparable performance on the tail class samples. Overall, \textit{RadGazeGen} improves the long-tailed classification performance on both the balanced and imbalanced test sets.

\subsubsection{Ablation Analysis.}
\label{ablation} 
Here, we discuss the image quality metrics for different components of \textit{RadGazeGen}, shown in Table \ref{tab:ablation}. 
% We report FID, SSIM and Peak Signal-to-Noise Ratio (PSNR) values for comparison.
We show the performance of Rad-CN, HVA-CN, and also individual radiomics maps with HVA-CN. For Rad-CN, we combined all the radiomics maps and do not use the HVA-CN component. For $\mathcal{R}_i+$HVA-CN, we use the different $\mathcal{R}_i$s, Canny, Sobel, GL, and segmentation masks individually with HVA-CN. We observe that \textit{RadGazeGen} outperforms all the individual components. Given that we use a similar experimental setup, it is interesting to note that Rad-CN performs slightly better than HVA-CN. Adding different $\mathcal{R}_i$s to HVA-CN improves the performance. This validates our initial hypothesis that adding radiomics filter maps to experts' gaze patterns will improve the quality of the generated medical images as they characterize complementary attributes (not captured by radiology reports). 
% \mb{The key takeaways are the composing radiomics feature maps with HVA inherently integrates textural patterns and disease patterns that are key components of medical imaging and hence improve the quality of image generation.}
\input{tables/table_3}

\begin{figure*}[t]
\centering
\includegraphics[height=9cm]{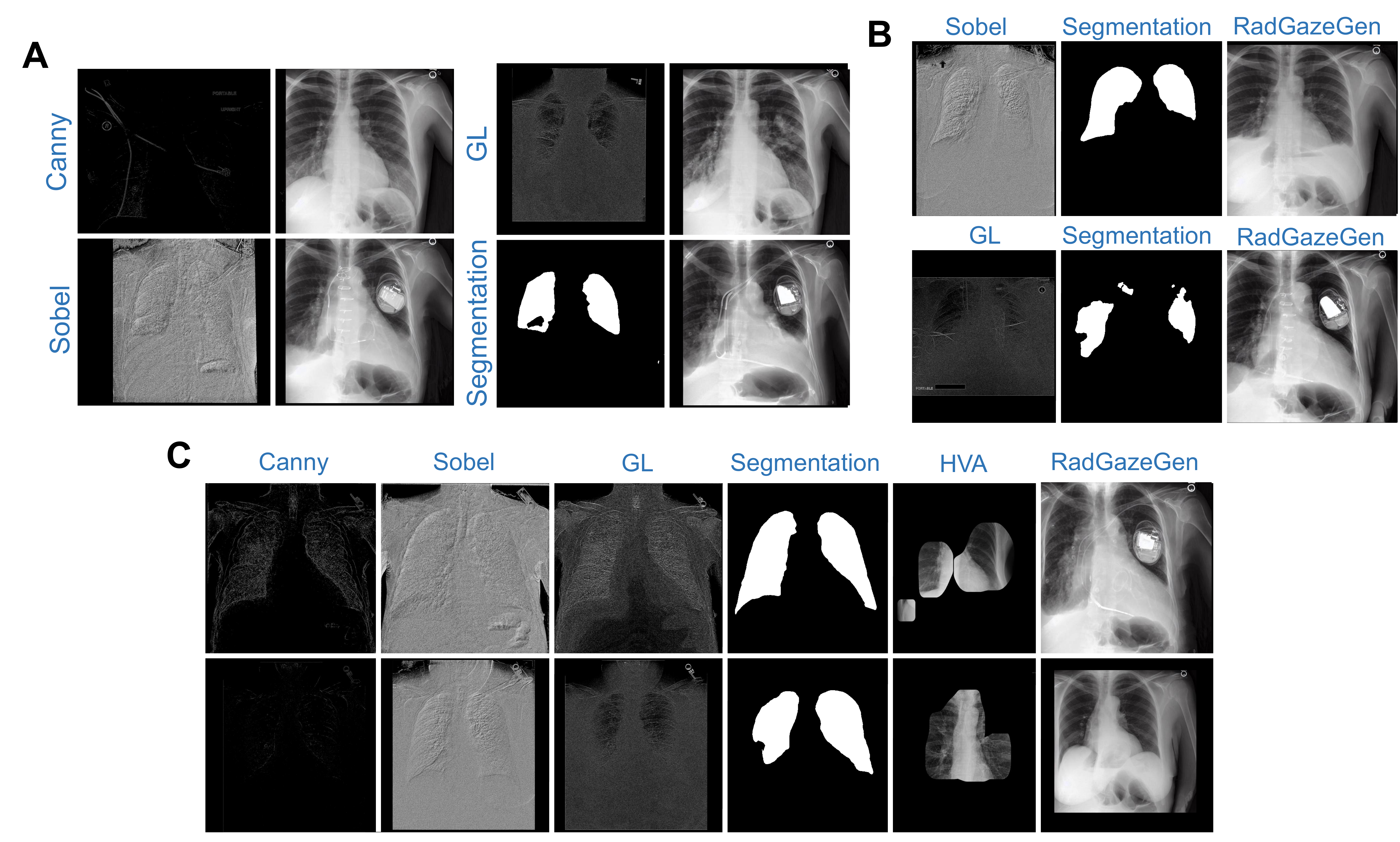}
\caption{\textbf{A.} A single control is used to generate CXRs, \textbf{B.} Multiple controls are used to generate CXRs, and \textbf{C.} $\mathcal{R}_i$ and $\mathcal{V}$ combined to generate CXRs.}
\label{fig:qualitative2}
\end{figure*}

\subsection{Qualitative Results} 
\begin{figure*}[t]
\centering
\includegraphics[height=12cm]{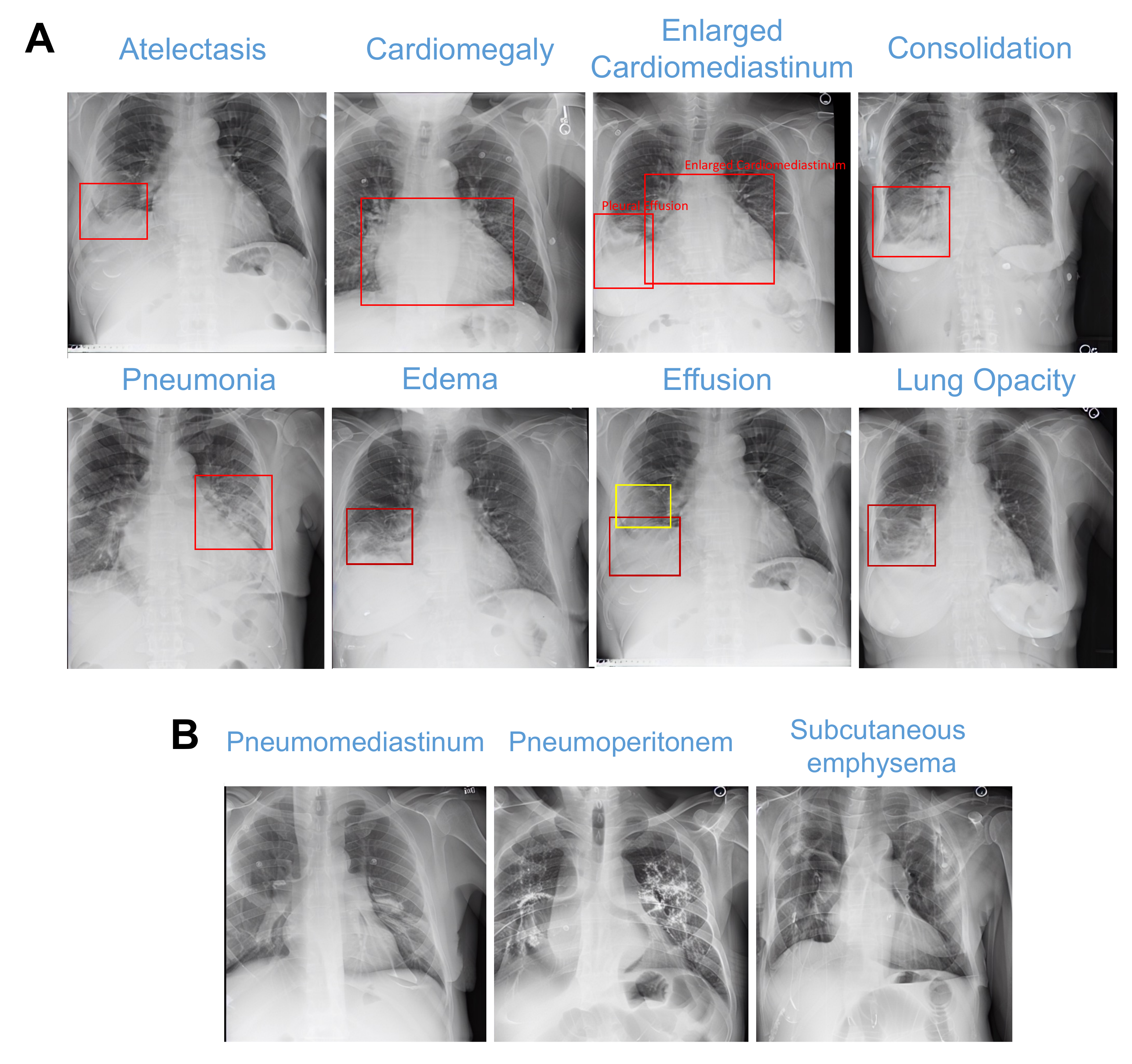}
\caption{\textbf{A.} CXRs generated for different pathologies from the CheXpert test dataset. We show the radiologist's annotation in \textcolor{red}{red bounding boxes} showing the location of the disease patterns. \textbf{B.} CXRs generated for the tail classes from the MIMIC-CXR-LT dataset.}
\label{fig:qualitative3}
\end{figure*}

% In this subsection, we discuss the qualitative results for our proposed method. 
Figure \ref{fig:qualitative1} shows the generated images for different disease pathologies like Consolidation, Effusion, Pneumothorax, etc.
% (more in Supplementary). 
We compare the images generated from \textit{RadGazeGen} with those from ControlNet and Multi-ControlNet, original images shown as a reference. We show the disease pattern annotations by a radiologist (7 years experience). We observe that \textit{RadGazeGen} not only generates high-quality images but also clinically accurate disease patterns. Whereas, ControlNet and MultiControlNet generate visually poor quality images as confirmed by a radiologist 
% (details in supplementary)
. The ground-truth text is shown in the first column with the corresponding disease highlighted in red (details in Supplementary). Figure \ref{fig:qualitative2} shows the generated images along with the controls. We observe that \textit{RadGazeGen} shows skillful composability of different controls. \textit{RadGazeGen} generates CXRs taking into consideration the textural patterns from radiomics feature maps with adherence to the segmentation masks and translates disease patterns and location from the HVA maps.
% In Figure \ref{fig:qualitative2}.A, we show the generated images when only one control is used. In Figure \ref{fig:qualitative2}.B, we we show the generated images when multiple controls are used in any combination. And, in Figure \ref{fig:qualitative2}.C, we show the \textit{GazeControlNet} generated images along with the controls. 
Figure \ref{fig:qualitative3} shows the \textit{RadGazeGen}-generated images for different disease pathologies using text and other controls from the CheXpert test and MIMIC-CXR-LT datasets. We show the bounding box annotations by the radiologist in red. We observe that \textit{RadGazeGen} generates clinically accurate disease patterns and hence improves disease classification performance.

\section{Discussion}
\label{discussion}

The findings presented in Section \ref{experiments_and_results} showcase the potential of \textit{RadGazeGen} in generating high-quality, clinically relevant CXR images from eye gaze patterns of radiologists and 
% textual descriptions and 
other control signals such as radiomics features. This discussion will delve deeper into the implications of these results, highlight the strengths and limitations of our approach, and suggest avenues for future research.

\subsection{Key Findings}

\textbf{Quality and Controllability.} The quantitative results, as detailed in Tables \ref{tab:quantitative1} and \ref{tab:quantitative2}, unequivocally demonstrate that \textit{RadGazeGen} outperforms state-of-the-art models such as T2I-Adapter, ControlNet, and Multi-ControlNet in generating images with superior FID scores, higher SSIM and mIoU. 
% The generated images are not only visually indistinguishable from real images but also maintain a high degree of structural and content fidelity provided by the input controls.
Eye gaze patterns of radiologists add clinically relevant information about disease pathology and fine-grained location to the generative models. 
% This improved controllability ensures that the generated images adhere closely to these clinically relevant controls
% the specified radiomics and HVA maps, 
% which is crucial for clinical applications where precision and accuracy are paramount. 
Furthermore, the integration of radiomics filter maps with eye gaze  enables the diffusion model to generate medical images with enhanced clinical fidelity, surpassing the capabilities of SDs. This multi-control composability is crucial for generating images that are not only visually realistic but also rich in clinically relevant details.

\textbf{Pulmonary Disease Classification Performance.} The classification results on the CheXpert dataset, summarized in Table \ref{tab:chexpert}, demonstrate that the images generated by \textit{RadGazeGen} are highly effective for disease classification tasks. The significant improvement in AUC scores for 7 out of 12 disease categories showcase the diagnostic utility of the synthetic images. This suggests that \textit{RadGazeGen} captures and preserves critical diagnostic features in the generated images, making them suitable for training and validating machine learning models.

\textbf{Long-Tailed Pulmonary Disease Classification Performance.} The results on the MIMIC-CXR-LT dataset, shown in Table \ref{tab:long_tailed}, highlight \textit{RadGazeGen}'s potential in addressing class imbalance. By generating additional images for underrepresented medium and tail classes, our model enhances the classification accuracy across these categories. This approach effectively mitigates the class imbalance problem, a common challenge in medical imaging datasets. The improvement in balanced and imbalanced test set performance demonstrates \textit{RadGazeGen}'s capability to generate diverse and representative synthetic data that can bolster the robustness of classification models.

% \subsection{Strengths of \textit{RadGazeGen}}

% \textbf{Composability of Controls.} A standout feature of \textit{RadGazeGen} is its adeptness at integrating multiple control signals. \mb{The integration of radiomics filter maps with eye gaze patterns enables the diffusion model to generate medical images with enhanced clinical fidelity, surpassing the capabilities of GANs.}
% % The combination of radiomics filter maps and eye gaze patterns Human Visual Attention (HVA) maps allows the model to 
% % capture both textural and pathological features of CXR images.
% This composability is crucial for generating images that are not only visually realistic but also rich in clinically relevant details. The ability to synthesize images that align with radiologists' interpretations and clinical annotations significantly enhances the model's utility in real-world clinical settings.

\textbf{Clinical Relevance.} The qualitative comparisons, depicted in Figures \ref{fig:qualitative1}, \ref{fig:qualitative2}, and \ref{fig:qualitative3}, further validate the clinical relevance of the images generated by \textit{RadGazeGen}. The generated images accurately reflect disease patterns and are consistent with the ground-truth annotations provided by experienced radiologists. This level of clinical fidelity is essential for applications in medical education, diagnostic support, and research. The model's ability to produce images that are both diagnostically and anatomically indistinguishable
from real images represents a significant advancement in medical image synthesis.

\subsection{Limitations}

Despite the promising results, our approach has certain limitations. The reliance on specific control signals, such as HVA maps, may restrict the generalizability of the model to other imaging modalities or clinical tasks where such data is not readily available. Additionally, the computational complexity associated with generating and training using multiple control signals requires substantial resources, which may limit the accessibility of our approach in resource-constrained environments. While our method is readily applicable to 3D imaging modalities, the computational cost of generating HVA maps increases significantly in 3D, thereby adding to the overall complexity of the proposed approach.

 % While our results demonstrate significant improvements, the generalizability of \textit{RadGazeGen} to diverse clinical populations and imaging conditions needs further validation. Ensuring robustness across varied datasets and clinical settings is crucial for real-world deployment.

\subsection{Future Directions}

% \textbf{Eye Gaze-Driven Innovations in Medical Image Generation.}
Incorporating eye gaze from radiologists into text-to-image diffusion models offers promising avenues for enhancing medical image generation. By aggregating gaze data from multiple experts, we can integrate diverse expertise into the generation process, potentially improving the accuracy and relevance of the produced images. 
% Additionally, extending this approach to 3D medical images allows for the generation of complex volumetric data, which can be invaluable in clinical diagnostics. 
Moreover, utilizing gaze patterns to generate images at different time points could provide insights into disease progression and treatment efficacy, offering a dynamic tool for personalized medicine and longitudinal studies. These advancements could significantly enhance the utility of generative models in medical imaging, contributing to better treatment decisions and patient outcomes.

Another promising direction is the extension of the \textit{RadGazeGen} framework to other medical imaging modalities, such as MRI or CT scans. This would involve adapting the model to handle 3D/4D radiomics features and control signals specific to these modalities. Expanding the applicability of our approach could significantly enhance its impact across a broader range of medical imaging tasks.

Finally, incorporating  additional control signals, such as patient demographics, clinical history, or genetic information, could further refine the specificity and accuracy of the generated images. This integration would allow for the creation of more contextually relevant medical images, enhancing their utility in precision medicine and patient-specific diagnostics.

\section{Conclusion}
In this work, we present \textit{RadGazeGen}, a novel  visual attention and radiomics-guided diffusion model for clinically accurate medical image generation. Unlike traditional text-conditioned image generation approaches which have suboptimal information on the structure, anatomy, and disease-relevant patterns in the medical images, our approach uses radiomics filter maps and eye gaze to generate better structured, anatomically accurate, and clinically relevant disease patterns. 
% \mb{}. 
The efficacy of our model is demonstrated in the quality of image generation and downstream disease classification problems. 
% In the future, we aim to increase our radiomics feature bank and further explore the clinical utility of the generated images. 
% In the future, we aim to develop clinician-in-the-loop medical image generation frameworks where the generated images are validated by medical experts' using eye gaze patterns.

\balance

\bibliographystyle{unsrt}  
\bibliography{references}

\end{document}

%% file: tables/table_1.tex
\begin{table}[tb]
  \caption{FID ($\downarrow$) metric is shown for different radiomics filter maps, segmentation masks, and HVA controls. We compare with different baselines. The best results are reported in \textbf{bold}. 
  }
  \label{tab:quantitative1}
  \centering
  \begin{tabular}{@{}llllll@{}}
    % \toprule
    \hline
    - & Canny & Sobel & GL & Segmentation & HVA\\
    % \midrule
    \hline
    % SD & 0.0 & 0.0 & 0.0 & 0.0 & 0.0\\
    % RoentGen & 0.0 & 0.0 & 0.0 & 0.0 & 0.0\\
    T2I-Adapter & \textbf{1.94} & \textbf{2.64} & 3.43 & 3.10 & 3.53\\
    ControlNet & 2.9338 & 2.9855 & 3.7689 & 2.6059 & 3.1240\\
    MultiControl & 3.1674 & 3.1707 & 2.9685 & 4.6870 & 3.0573\\
    % Uni-ControlNet & 0.0 & 0.0 & 0.0 & 0.0 & 0.0\\
    \hline
    RadGazeGen & 2.8697 & 2.9065 & \textbf{2.9111} & \textbf{2.5820} & \textbf{3.0021}\\
  % \bottomrule
  \hline
  \end{tabular}
\end{table}

\begin{table}[tb]
  \caption{Assessment of contollability using SSIM($\uparrow$), mIoU($\uparrow$), and CLIP-score ($\uparrow$) for different controls. The best results are shown in \textbf{bold}.}
  \label{tab:quantitative2}
  \centering
  \begin{tabular}{@{}llllll@{}}
    % \toprule
    \hline
    - & \makecell{Canny\\(SSIM)} & \makecell{Sobel\\(SSIM)} & \makecell{GL\\(SSIM)} & \makecell{Segment.\\(mIoU)} & \makecell{Style/Content\\(CLIP Score)}\\
    % \midrule
    \hline
    SD & 0.3091 & \textbf{0.0080} & 0.0090 & 0.3111 & 27.6936\\
    RoentGen & 0.1942 & 0.0079 & 0.0087 & \textbf{0.3689} & 27.4786\\
    % T2I-Adapter & 0.5041 & 0.0074 & 0.0095 & \textbf{0.4715} & 27.9619\\
    T2I-Adapter & 0.53 & 0.0079 & 0.0087 & 0.06 & 27.62\\
    ControlNet & 0.4526 & 0.0078 & 0.0095 & 0.3599 & 27.7032\\
    MultiControl & 0.5816 & 0.0077 & 0.0090 & 0.1056 & 22.2958\\
    % Uni-ControlNet & 0.0 & 0.0 & 0.0 & 0.0 & 0.0\\
    \hline
    RadGazeGen & \textbf{0.7045} & 0.0076 & \textbf{0.0101} & 0.3555 & \textbf{27.7757}\\
  % \bottomrule
  \hline
  \end{tabular}
\end{table}

%% file: tables/table_2.tex
% \begin{table}[tb]
%   \caption{Quantitative Results
%   }
%   \label{tab:headings}
%   \centering
%   \begin{tabular}{@{}llllll@{}}
%     \toprule
%     - & Canny & Sobel & GL & Segmentation & Style/Content\\
%     \midrule
%     T2I-Adapter & 0.0 & 0.0 & 0.0 & 0.0 & 0.0\\
%     ControlNet & 0.0 & 0.0 & 0.0 & 0.0 & 0.0\\
%     MultiControl & 0.0 & 0.0 & 0.0 & 0.0 & 0.0\\
%     % Uni-ControlNet & 0.0 & 0.0 & 0.0 & 0.0 & 0.0\\
%     \hline
%     Gaze-ControlNet & 0.0 & 0.0 & 0.0 & 0.0 & 0.0\\
%   \bottomrule
%   \end{tabular}
% \end{table}

%\begin{table}[tb]
%  \caption{Classification - NIH
%  }
%  \label{tab:headings}
%  \centering
%  \begin{tabular}{@{}llllllllllllll@{}}
%    \toprule
%    - & Ate. & Cmg. & Cns. & Edm. & Les. & Opa. & Eff. & PTh & Pne. & PTX & Fra. & Enl.\\
%    \midrule
%    % Original & 0.0 & 0.0 & 0.0 & 0.0 & 0.0 & 0.0 & 0.0 & 0.0 & 0.0 & 0.0 & 0.0 & 0.0\\
%    SD & 0.0 & 0.0 & 0.0 & 0.0 & 0.0 & 0.0 & 0.0 & 0.0 & 0.0 & 0.0 & 0.0 & 0.0\\
%    RoentGen & 0.0 & 0.0 & 0.0 & 0.0 & 0.0 & 0.0 & 0.0 & 0.0 & 0.0 & 0.0 & 0.0 & 0.0\\
%    % T2I-Adapter & 0.0 & 0.0 & 0.0 & 0.0 & 0.0 & 0.0 & 0.0 & 0.0 & 0.0 & 0.0 & 0.0 & 0.0\\
%    ControlNet & 0.0 & 0.0 & 0.0 & 0.0 & 0.0 & 0.0 & 0.0 & 0.0 & 0.0 & 0.0 & 0.0 & 0.0\\
%    MultiControl & 0.0 & 0.0 & 0.0 & 0.0 & 0.0 & 0.0 & 0.0 & 0.0 & 0.0 & 0.0 & 0.0 & 0.0\\
%    % Uni-ControlNet & 0.0 & 0.0 & 0.0 & 0.0 & 0.0\\
%    \hline
%    GazeControlNet & 0.0 & 0.0 & 0.0 & 0.0 & 0.0 & 0.0 & 0.0 & 0.0 & 0.0 & 0.0 & 0.0 & 0.0\\
%  \bottomrule
%  \end{tabular}
%\end{table}

\begin{table*}[tb]
  \caption{Classification performance on generated images using text and other controls from the CheXpert test dataset. AUC($\uparrow$) is reported for different disease pathologies. The best results are shown in \textbf{bold}.
  }
  \label{tab:chexpert}
  \centering
  \scalebox{0.8}{
  \begin{tabular}{@{}lllllllllllllll@{}}
    % \toprule
    \hline
    - & Ate. & Cmg. & Cns. & Edm. & Les. & Opa. & Eff. & PTh & Pne. & PTX & Fra. & Enl. & Avg.\\
    % \midrule
    \hline
    Original$^*$ & 81.95 & 86.45 & 89.35 & 86.85 & 78.76 & 90.31 & 89.77 & 85.04 & 90.09 & 76.49 & 58.26 & 78.11 & 82.62\\
    SD & 72.57 & \textbf{73.64} & 73.77 & \textbf{75.28} & 64.24 & 83.32 & 74.59 & \textbf{68.07} & 59.24 & 58.36 & \textbf{71.70} & \textbf{78.67} & 71.12\\
    % RoentGen & 76.19 & 77.17 & 85.08 & 80.74 & 60.31 & 84.32 & 84.35 & 63.11 & 74.03 & 64.54 & 44.44 & 78.45\\
    % % \hline
    % \hdashline
    % % T2I-Adapter & 0.0 & 0.0 & 0.0 & 0.0 & 0.0 & 0.0 & 0.0 & 0.0 & 0.0 & 0.0 & 0.0 & 0.0\\
    % ControlNet & 71.42 & 65.42 & 71.61 & 63.77 & \textbf{67.89} & 80.30 & 71.31 & 51.44 & 70.55 & 58.31 & 40.63 & 73.35\\
    % MultiControlNet & 70.29 & 65.11 & 69.57 & 65.13 & 63.41 & 78.96 & 70.14 & 48.63 & 68.86 & 51.93 & 52.06 & 71.89\\
    % Uni-ControlNet & 0.0 & 0.0 & 0.0 & 0.0 & 0.0\\
    T2I-Adapter & 47.8 & 47.2 & 64.2 & 62.3 & 36.3 & 63.0 & 48.0 & 58.6 & 60.64 & \textbf{66.2} & 57.2 & 53.76 & 55.43\\
    \hline
    % \hline
    RadGazeGen & \textbf{80.71} & 70.34 & \textbf{77.67} & 73.47 & \textbf{66.57} & \textbf{88.20} & \textbf{78.31} & 53.29 & \textbf{77.46} & 62.74 & 52.32 & 74.90 & \textbf{71.33}\\
  % \bottomrule
  \hline
  \multicolumn{14}{l}{\tiny *The first row reports the model's performance on the real images from the CheXpert test dataset.} \\
  \end{tabular}
  % \mb{The first row reports the model's performance on the real images from the CheXpert test dataset.}
  }
  % \myfootnotetext{The first row reports the model's performance on the real images from the CheXpert test dataset.}
% \footnotetext[1]{The first row reports the model's performance on the real images from the CheXpert test dataset.}
\end{table*}
% \footnotesize{aaa}

% \begin{table}[tb]
%   \caption{Long-Tailed Classification - NIH-CXR-LT
%   }
%   \label{tab:headings}
%   \centering
%   \begin{tabular}{@{}lllllll@{}}
%     \toprule
%     - & Head & Medium & Tail & Avg & F1 & bAcc\\
%     \midrule
%     Softmax & 0.419 & 0.056 & 0.017 & 0.164 & 0.131 & 0.115\\
%     Focal & 0.362 & 0.056 & 0.042 & 0.153 & 0.142 & 0.122\\
%     LDAM & 0.410 & 0.133 & 0.142 & 0.228 & 0.173 & 0.178\\
%     RW LDAM-DRW & 0.410 & 0.367 & 0.308 & 0.362 & 0.127 & 0.289\\
%     Decoupling-cRT & 0.433 & 0.374 & 0.300 & 0.369 & 0.138 & 0.294\\
%     % SD & 0.0 & 0.0 & 0.0 & 0.0 & 0.0\\
%     % T2I-Adapter & 0.0 & 0.0 & 0.0 & 0.0 & 0.0\\
%     ControlNet & 0.0 & 0.0 & 0.0 & 0.0 & 0.0\\
%     MultiControl & 0.0 & 0.0 & 0.0 & 0.0 & 0.0\\
%     % Uni-ControlNet & 0.0 & 0.0 & 0.0 & 0.0 & 0.0\\
%     \hline
%     GazeControlNet & 0.0 & 0.0 & 0.0 & 0.0 & 0.0\\
%   \bottomrule
%   \end{tabular}
% \end{table}

\begin{table}[tb]
  \caption{Long-Tailed Classification on the MIMIC-CXR-LT balanced test ($n=570$) and the test set ($n=23550$). Accuracy is reported for the individual head, medium, and tail classes and average of all the classes (Avg) is reported for the balanced set. For the imbalanced set, Macro-F1 (mF1) and balanced accuracy (bAcc) are reported.
  }
  \label{tab:long_tailed}
  \centering
  \begin{tabular}{@{}lllllll@{}}
    % \toprule
    \hline
     - & \multicolumn{4}{c}{\bfseries Balanced test} & \multicolumn{2}{c}{\bfseries Test}\\
     \hline
    - & Head & Medium & Tail & Avg & mF1 & bAcc\\
    % \midrule
    \hline
    % Softmax$^\triangle$ & 0.503 & 0.039$^*$ & 0.022$^*$ & 0.188$^*$ & 0.183$^*$ & 0.169\\     
    % Focal & 0.477 & 0.044 & 0.022 &  0.181 & 0.182 & 0.172\\     
    % LDAM & 0.497 & 0.0 & 0.0 & 0.166 & 0.172 & 0.165\\
    % RW LDAM-DRW & 0.447 & 0.256 & 0.311 & 0.338 & 0.177 & 0.275\\
    % Decoupling-cRT & 0.490 & 0.306 & 0.367 & 0.387 & 0.170 & 0.296\\
    % % SD & 0.0 & 0.0 & 0.0 & 0.0 & 0.0\\
    % % T2I-Adapter & 0.0 & 0.0 & 0.0 & 0.0 & 0.0\\
    % \hdashline
    \makecell{ControlNet\\(Canny)} & 0.497 & 0.016 & 0.011 & 0.175 & 0.174 & 0.169\\
    \makecell{MultiControlNet\\(Canny+Seg)}  & 0.504 & 0.011 & \textbf{0.022} & 0.179 & 0.181 & 0.171\\
    % Uni-ControlNet & 0.0 & 0.0 & 0.0 & 0.0 & 0.0\\
    \hline
    RadGazeGen & \textbf{0.506} & \textbf{0.027} & \textbf{0.022} & \textbf{0.185} & \textbf{0.182} & \textbf{0.178}\\
    % GazeControlNet$^{**}$ & 0.0 & 0.0 & 0.0 & 0.0 & 0.0\\
  % \bottomrule
  \hline
  \end{tabular}
\end{table}

%% file: tables/table_3.tex
\begin{table}[tb]
  \caption{\textbf{Ablation results.}  FID($\downarrow$), SSIM($\uparrow$), and PSNR($\uparrow$) for different components of the \textit{RadGazeGen} architecture.
  }
  \label{tab:ablation}
  \centering
  \begin{tabular}{@{}llll@{}}
    % \toprule
    \hline
    - & FID($\downarrow$) & SSIM($\uparrow$) & PSNR($\uparrow$)\\
    % \midrule
    \hline
    Rad-CN & 3.0679 & 0.6373 & \textbf{10.6014}\\ %
    HVA-CN & 3.0511 & 0.6316 & 10.1067\\ %
    Canny+HVA-CN & 3.0554 & 0.6373 & 10.2443\\ %
    GL+HVA-CN & 3.0552 & 0.6368 & 10.2068\\ %
    Segmentation+HVA-CN & 3.0485 & 0.6362 & 10.2484\\ %
    \hline
    RadGazeGen & \textbf{2.9912} & \textbf{0.6510} & \textbf{10.6014}\\ %
  % \bottomrule
  \hline
  \end{tabular}
\end{table}